\documentclass[10pt,twocolumn,letterpaper]{article}

\usepackage{wacv}
\usepackage{times}
\usepackage{epsfig}
\usepackage{graphicx}
\usepackage{amsmath}
\usepackage{amssymb}
\usepackage{multirow}
\usepackage{url}

\wacvfinalcopy % *** Uncomment this line for the final submission

\def\wacvPaperID{320} % *** Enter the wacv Paper ID here

\ifwacvfinal\fi
\begin{document}

%%%%%%%%% TITLE
\title{Decoupled Learning for Conditional Adversarial Networks}

% Authors at the same institution
%\author{First Author \hspace{2cm} Second Author \\
% Authors at different institutions
\author{Zhifei Zhang, Yang Song, and Hairong Qi \\
University of Tennessee\\
{\tt\small \{zzhang61, ysong18, hqi\}@utk.edu}
}

\maketitle
\ifwacvfinal\thispagestyle{empty}\fi

%%%%%%%%% ABSTRACT
\begin{abstract}
   Incorporating encoding-decoding nets with adversarial nets has been widely adopted in image generation tasks. We observe that the state-of-the-art achievements were obtained by carefully balancing the reconstruction loss and adversarial loss, and such balance shifts with different network structures, datasets, and training strategies. Empirical studies have demonstrated that an inappropriate weight between the two losses may cause instability, and it is tricky to search for the optimal setting, especially when lacking prior knowledge on the data and network.
   This paper gives the first attempt to relax the need of manual balancing by proposing the concept of \textit{decoupled learning}, where a novel network structure is designed that explicitly disentangles the backpropagation paths of the two losses. 
   In existing works, the encoding-decoding nets and GANs are integrated by sharing weights on the generator/decoder, thus the two losses are backpropagated to the generator/decoder simultaneously, where a weighting factor is needed to balance the interaction between the two losses. The decoupled learning avoids the interaction and thus removes the requirement of the weighting factor, essentially improving the generalization capacity of the designed model to different applications. The decoupled learning framework could be easily adapted to most existing encoding-decoding-based generative networks and achieve competitive performance without the need of weight adjustment. 
   Experimental results demonstrate the effectiveness, robustness, and generality of the proposed method. The other contribution of the paper is the design of a new evaluation metric to measure the image quality of generative models. We propose the so-called \textit{normalized relative discriminative score} (NRDS), which introduces the idea of relative comparison, rather than providing absolute estimates like existing metrics. The demo code is available\footnote{\color{blue}\url{https://github.com/ZZUTK/Decoupled-Learning-Conditional-GAN}}.
\end{abstract}

%%%%%%%%% BODY TEXT
\section{Introduction}
\label{sec:introduction}
Generative adversarial networks (GANs)~\cite{goodfellow2014generative} is an adversarial framework that generates images from noise while preserving high fidelity. However, generating random images from noise doesn't meet the requirements in many real applications, e.g., image inpainting~\cite{pathak2016context}, image transformation~\cite{isola2016image,song2017recursive}, image manipulation~\cite{zhang2017age,zhu2016generative}, etc. To overcome this problem, recent works like ~\cite{reed2016generative,nguyen2016synthesizing} concatenate additional features generated by an encoder or certain extractor to the random noise or directly replace the noise by the features.
In most recent practices, the encoding-decoding networks (ED), e.g., VAE~\cite{kingma2013auto}, AAE~\cite{makhzani2015adversarial}, Autoencoder~\cite{liou2014autoencoder}, etc., have been the popular structure to be incorporated with GANs~\cite{goodfellow2014generative} for image-conditional modeling, where the encoder extracts features, which are then fed to the decoder/generator to generate the target images. The encoding-decoding network tends to yield blurry images. Incorporating a discriminator, as empirically demonstrated in many works~\cite{larsen2015autoencoding,isola2016image,ledig2016photo,zhang2017age,liu2017unsupervised,zhu2017unpaired,zhang2017gans}, effectively increases the quality (i.e., reality and resolution) of generated images from the encoding-decoding networks. In recent two years, the adversarial loss has become a common regularizer for boosting image quality, especially in image generation tasks.

\begin{figure}[ht]
	\centering
	\includegraphics[width=.9\columnwidth]{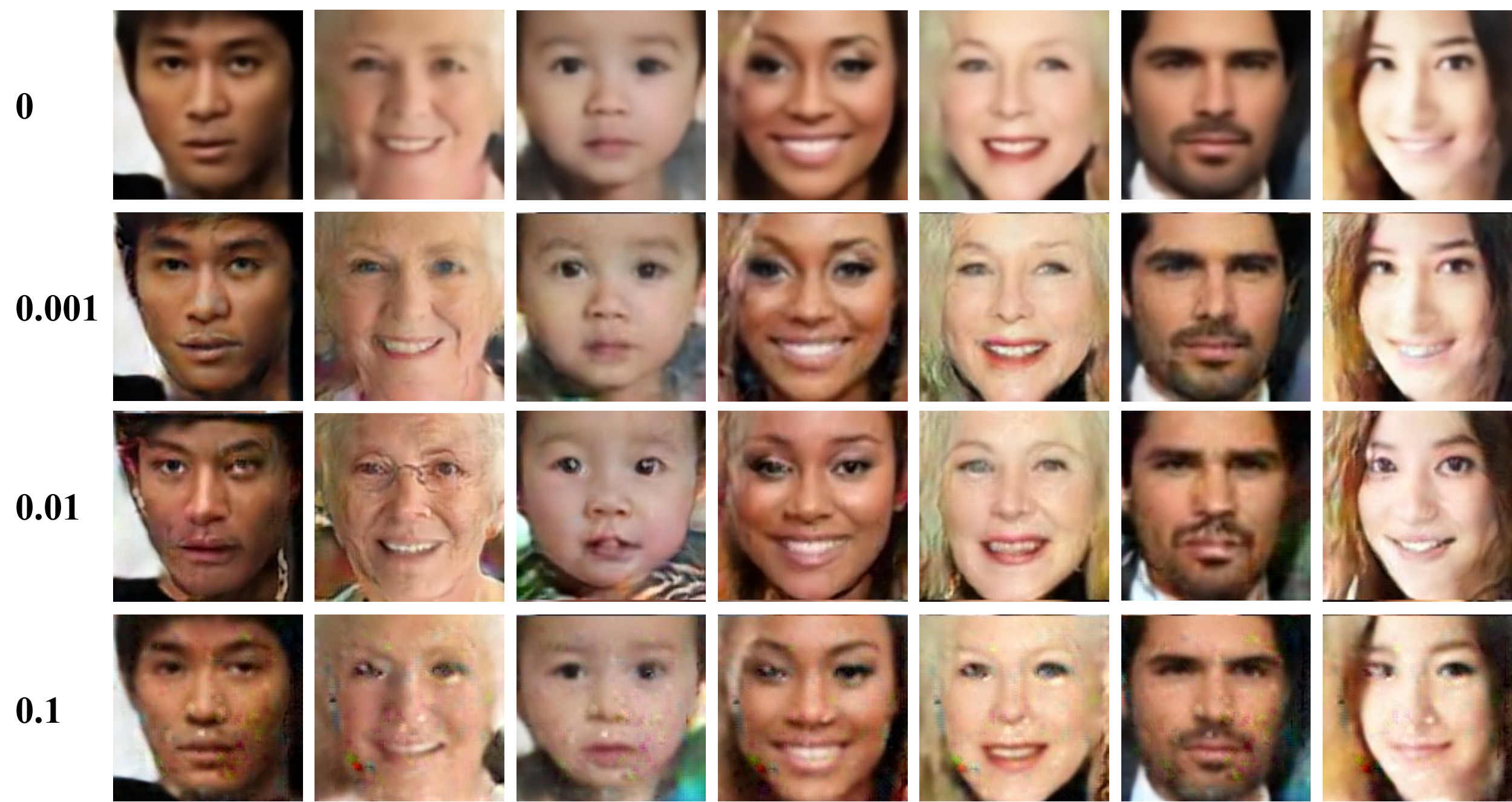}
	\caption{Results generated from the coupled network of ED and GAN. Please zoom in for more details. The weight of reconstruction loss is 1, and the weight of adversarial loss is on the left.}
	\label{fig:artifacts}
\end{figure}

In existing works that incorporate the encoding-decoding networks (ED) with GANs, the decoder of ED and generator of GAN share the same network and parameters, thus the reconstruction loss (from ED) and the adversarial loss (from discriminator) are both imposed on the decoder/generator. Although ED is known to be stable in training, and many alternatives of GANs, e.g., DCGAN~\cite{radford2015unsupervised}, WGAN~\cite{arjovsky2017wasserstein}, LSGAN~\cite{mao2016least}, etc., have stabilized the training of GANs, coupling the reconstruction loss and the adversarial loss by making them interact with each other may yield unstable results, e.g., introducing artifacts as shown in Fig.~\ref{fig:artifacts}.
We observe the increased details of generated images as compared to the image generated from ED only (the top row in Fig.~\ref{fig:artifacts} where the weight of adversarial loss is 0). However, we also observe the obvious artifacts introduced by adding the adversarial loss (e.g., the 1st, 2nd faces with weights $0.01$ and $0.1$). A higher weight on the adversarial loss preserves richer details in generated images but suffering higher risk of introducing significant artifacts or even causing instability, while a lower weight on the adversarial loss would not effectively boost the image fidelity. Generally, the trade-off between the two losses needs to be carefully tuned, otherwise, the generated images may present significant artifacts, e.g., stripe, spots, or anything visually unrealistic.   

Existing works generally arrive at an appropriate weight between the two losses by conducting extensive empirical study; and yet this weight 
%achieve the peak performance based regardless of the difficulty of balancing the reconstruction loss (from ED) and adversarial loss (from GAN). Mostly, an appropriate weight between the two losses was reported without reasoning, and the appropriate weight 
may vary with different network structures or different datasets used. 
%A lot of empirical studies have demonstrated that the weight setting between the two losses could be difficult because an inappropriate setting causes unstable results, and the changing of network structure and dataset may significantly change the appropriate setting. Until today, there isn't a convincing clue that could guide the weight setting. For example, given the network in~\cite{isola2016image} and the corresponding dataset, how would you set the weight between the two losses during training if you haven't read that paper and don't have any experience in the deep network training? Maybe, you have to randomly try or exhaustively search in worst. This paper will relax the unfounded and elusive weight setting.   

In this paper, we give the first attempt to relax the manual balancing between the two losses by proposing a novel \textit{decoupled learning} structure. Moving away from the traditional routine of incorporating the ED and GAN, decoupled learning explicitly disentangles the two networks, avoiding interaction between them. To %terminologically distinguish the proposed decoupled learning from existing works, 
make the presentation easy to follow, we denote the coupled structure used in existing works as ED+GAN\footnote{\tiny The coupled structures used in existing works are denoted as ED+GAN because they add the effects of ED and GAN together during training.}, and the proposed method  as ED//GAN\footnote{\tiny The proposed decoupled learning is denoted as ED//GAN, indicating that the effect from ED and GAN are learned/propagated separately through the two networks.}. The contributions of this paper could be summarized from the following three aspects:      
\begin{itemize}
	\item We propose the decoupled learning (ED//GAN) to tackle the ubiquitous but often neglected problem in the widely adopted ED+GAN structure that removes the need for manual  balancing between the reconstruction loss and adversarial loss. To the best of our knowledge, this is the first attempt to deal with this issue.
	\item Based on the proposed decoupled learning (ED//GAN), we further observe its merit in visualizing the boosting effect of adversarial learning. Although many empirical studies  demonstrated the effect of GAN in the visual perspective, few of them could %straightforwardly
	demonstrate how GAN sharpens the blurry output from ED, e.g., what kinds of edges and textures could be captured by GAN but missed by ED.   
	\item %As the lack of convincing evaluation metric for generative models, 
	Moving away from providing absolute performance metrics like existing works, we design the \textit{normalized relative discriminative score} (NRDS) that provides relative estimates of the models in comparison. After all, the purpose of model evaluation is mostly to rank their performance; therefore, many times, absolute measurements are unnecessary. In essence, NRDS aims to illustrate whether one model is better or worse than another, which is more practical to arrive at a reliable estimate.      
\end{itemize}

\section{Decoupled Learning}  
\label{sec:decoupled_learning}  
In the widely used network structure ED+GAN, ED appears to generate smooth and blurry results due to minimization of pixel-wise average of possible solutions in the pixel space, while GAN drives results towards the natural image manifold producing perceptually more convincing solutions. Incorporating the two parts as in existing works causes competition between the two networks, and when the balance point is not appropriately chosen, bad solutions might result causing artifacts in the generated images. Many empirical studies have demonstrated that it does not necessarily boost the image quality by topping a GAN to ED. We aim to avoid such competition by training ED and GAN in a relatively independent manner -- we preserve the structures of ED and GAN without sharing parameters, as compared to existing works where the parameters of decoder in ED and generator in GAN are shared. The independent network design explicitly decouples the interaction between ED and GAN, but still follows the classic objective functions --- the reconstruction loss and minimax game for ED and GAN, respectively. Thus, any existing work based on ED+GAN can be easily adapted to the proposed structure without significantly changing their objectives, meanwhile gaining the benefit of not having to find a balance between ED and GAN. 
\subsection{Difference between ED+GAN and ED//GAN} 
\begin{figure}[ht]
	\centering
	\includegraphics[width=.9\columnwidth]{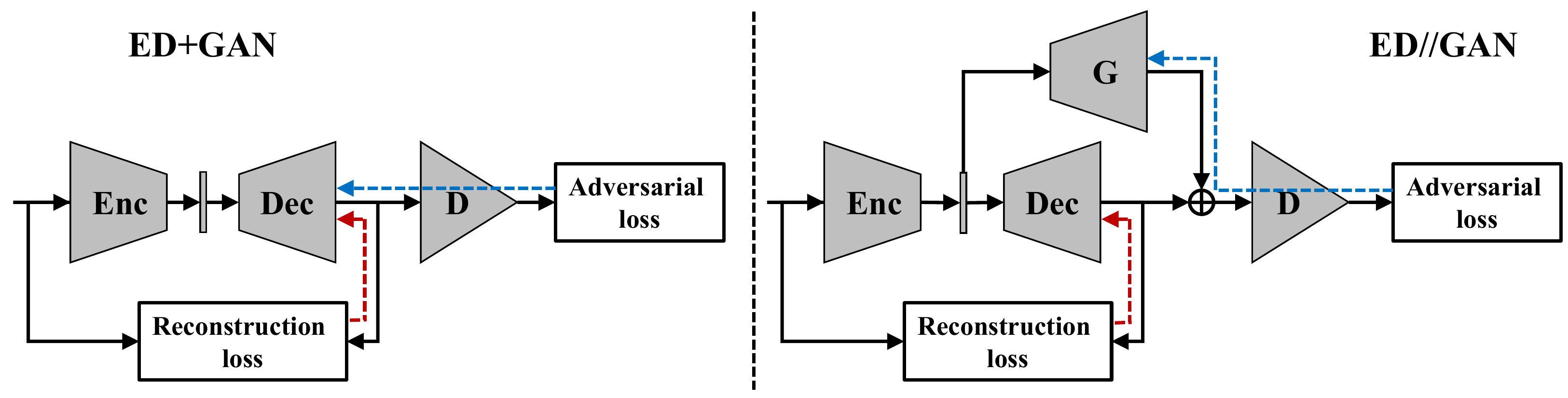}
	\caption{Comparison between ED+GAN and ED//GAN. Left: the existing ED+GAN. Right: the proposed ED//GAN, i.e., decoupled learning. Enc and Dec are the encoder and decoder networks, and G and D are the generator and discriminator, respectively. Solid black arrows denote the feedforward path, and dashed arrows in red and blue indicate backpropagation of the reconstruction loss and the adversarial loss, respectively.}
	\label{fig:demo}
\end{figure}

Compared to ED+GAN, the uniqueness of the proposed ED//GAN lies in the two decoupled backpropagation paths where the reconstruction and adversarial losses are backpropagated to separate networks, instead of imposing both losses to generator/decoder as done in ED+GAN. Fig.~\ref{fig:demo} illustrates the major difference between ED+GAN and ED//GAN.   

In ED+GAN, both reconstruction loss and adversarial loss are backpropagated to Dec, and the general objective could be written as
\begin{equation}
	\underset{Enc,Dec,D}{\min}\;\mathcal{L}_{const} + \lambda\mathcal{L}_{adv}, %\;\;\;\text{ or }\;\;\;  %\underset{Enc,Dec,D}{\min}\;\lambda\mathcal{L}_{const} + \mathcal{L}_{adv},
	\label{eq:obj_old}
\end{equation}
where $\mathcal{L}_{const}$ and $\mathcal{L}_{adv}$ denote the reconstruction and adversarial losses, respectively. The parameter $\lambda$ is the weight to balance the two losses. %During the training, $\mathcal{L}_{const}$ updates Enc and Dec, and $\mathcal{L}_{adv}$ updates D and Dec (or together with Enc). Dec will be simultaneously updated by the two losses. 

In ED//GAN, we are no longer in need of the weight $\lambda$ because the backpropagation from two losses are along different paths without interaction. Then, the general objective for ED//GAN becomes
\begin{align}
	~ & \underset{Enc,Dec,G,D}{\min}\;\mathcal{L}_{const} + \mathcal{L}_{adv}\\
	= & \underset{Enc,Dec}{\min}\;\mathcal{L}_{const} + \underset{G,D}{\min}\;\mathcal{L}_{adv}.
	\label{eq:obj_new}
\end{align}

\subsection{The General Framework}
\begin{figure}[ht]
	\centering
	\includegraphics[width=\columnwidth]{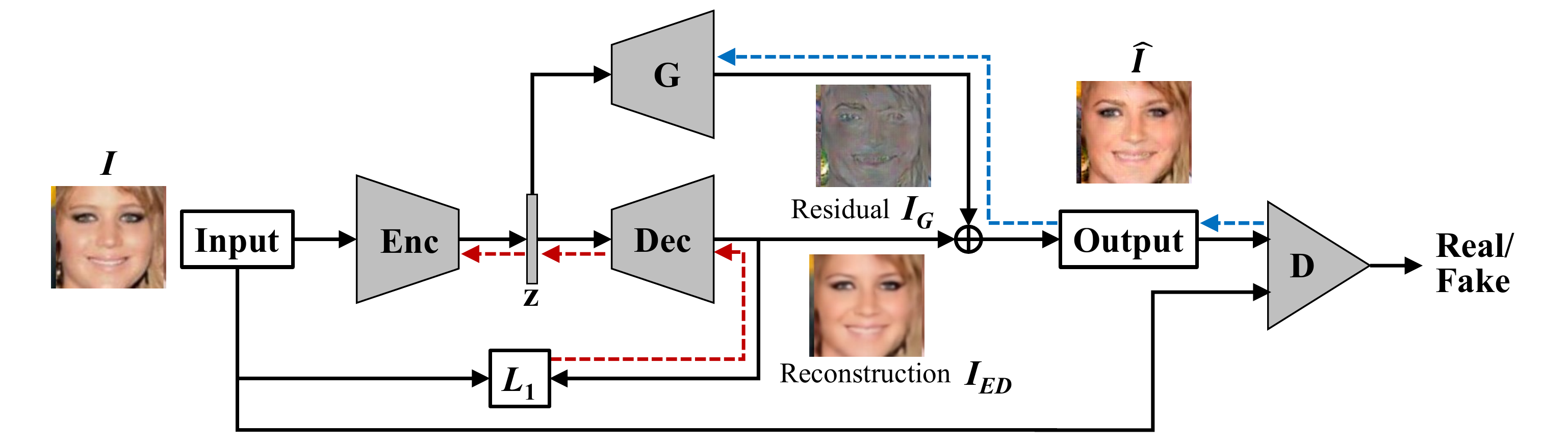}
	\caption{The flow of proposed decoupled learning, i.e., ED//GAN. $L_1$ indicates the pixel-level $\ell_1$-norm. Solid black arrows denote the feedforward path, and dashed arrows in red and blue indicate the backpropagation from reconstruction loss ($L_1$) and adversarial loss (from D), respectively. The reconstructed image $I_{ED}$ is generated from the decoder (Dec), and the residual $I_G$ is generated from the generator (G). G and Dec share the latent variable $z$ derived from the encoder (Enc). The final output image $\hat{I}$ is obtained through pixel-wised adding the two generated images $I_G$ and $I_{ED}$, as indicated by the $\bigoplus$ marker.}
	\label{fig:flow}
\end{figure}

The general framework of the proposed decoupled learning (ED//GAN) is detailed in Fig.~\ref{fig:flow}, incorporating the encoding-decoding network (ED) with GAN (i.e., D and G) in a decoupled manner, i.e., G and Dec are trained separately corresponding to the adversarial loss and reconstruction loss, respectively.  
Assuming the input image $I\in\mathbb{R}^{H\times W\times C}$, where $H$, $W$, and $C$ denote the height, width, and the number of channels, respectively.
ED (i.e., Enc and Dec) is trained independently from GAN (i.e., G and D), and the reconstructed image from ED is $I_{ED}$, which is a blurred version of the input image $I$. The generator G, together with the discriminator D, learns $I_G$ which is added to $I_{ED}$ to yield the final output image $\hat{I}$. Since $I \approx \hat{I} = I_{ED}+I_G$, the generated images from G is actually the residual map between $I_{ED}$ and $\hat{I}$. Assuming $\hat{I}$ is close to the real image $I$, then $I_G$ would illustrate how adversarial learning increases the resolution and photo-realism of a blurry image. Generally, $I_G$ contains details, e.g., edges and texture, and specifically, wrinkles and edges of the eyes and mouth in face images. Therefore, a byproduct of the decoupled learning is that it provides a direct mechanism to conveniently illustrate how the adversarial learning boosts the performance of ED.

In the proposed ED//GAN framework, the gradient derived from the reconstruction loss and adversarial loss are directed in separated paths without any interaction, avoiding the competition between reconstruction and adversarial effects which may cause instability as discussed in the introduction. 
G serves as the subsequent processing block of ED, recovering details missed by the output from ED. The G and Dec share the latent variable because of the correspondence between the blurry image $I_{EG}$ and the corresponding recoverable details $I_G$. %For example, the details mainly contain the wrinkle, edges of eyes and mouth, etc, which vary conditioned on the feature of faces. Therefore, we assume that $I_{EG}$ and $I_G$ should share the high-level features.  

\subsection{Training of the ED//GAN}
The proposed decoupled learning can be %graphically
divided into two parts: 1) reconstruction learning of Enc and Dec and 2) adversarial learning of G and D. Enc and Dec (i.e., ED) are trained independently of G and D (i.e., GAN), updated through the $\ell_1-$norm in pixel level as shown by the red dashed arrow in Fig.~\ref{fig:flow}. G and D are trained by the original objective of GAN, and G is only updated by the adversarial loss as indicated by the blue dashed arrow. The final output image is obtained by pixel-wise summation of the outputs from G and Dec.  

\subsubsection{Reconstruction Learning}
The encoding-decoding network (ED) aims to minimize the pixel-level error between the input image $I$ and reconstructed image $I_{ED}$. The training of ED is well known to be stable, and ED could be any structures specifically designed for any applications, e.g., U-Net~\cite{isola2016image} or conditional network~\cite{zhang2017age} with/without batch normalization. Most works that adopted batch normalization to enhance stability of the ED+GAN structure may bring a few unfortunate side effects~\cite{goodfellow2016nips} and hurt the diversity~\cite{zhang2017age} of generated images. With the proposed ED//GAN, however, batch normalization becomes unnecessary because the training of ED is isolated from that of GAN, and ED itself could be stable without batch normalization. 
The reconstruction loss of the ED part can be expressed by 
\begin{align}
	\mathcal{L}_{const}(Enc,Dec) = & \| I - Dec(Enc(I)) \|_1 \\
	= & \| I - Dec(z) \|_1 \\
	= & \| I - I_{ED} \|_1,
	%\label{eq:l_const}
\end{align} 
where $Enc$ and $Dec$ indicate the functions of encoder and decoder, respectively. The latent variable derived from $Enc$ is denoted by $z$. $\|\cdot\|_1$ indicates $\ell_1$-norm in pixel level. More general, the latent variable $z$ could be constrained to certain prior distribution (e.g., Gaussian distribution or uniform distribution) to achieve generative ability like in VAE~\cite{kingma2013auto} and AAE~\cite{makhzani2015adversarial}.

\subsubsection{Adversarial Learning}
%In GANs, two competitive components, namely generator (G) and discriminator (D), are trained simultaneously through a minimax game. The D aims to discriminate the real samples (input images) from faked ones (generated by G), while G aims to fool D.  
In the proposed ED//GAN, GAN works differently from the vanilla GAN in two aspects: 1) The inputs of G are features of the input image (sharing the latent variable $z$ with Dec) rather than the random noise. 2) The fake samples fed to D are not directly generated by G. Instead, they are conditioned on the output from Dec. Therefore, the losses of GAN can be expressed as
%For simplicity, we write the losses of G and D separately,
\begin{align}
	\mathcal{L}_{adv}(D)=&\mathbb{E}\left[ \log \left(1-D(I)\right) \right] + \mathbb{E}\left[ \log D\left(I_{ED}+I_G\right) \right)],\label{eq:l_adv_d}\\
	\mathcal{L}_{adv}(G)=&\mathbb{E}\left[ \log \left( 1 - D\left(I_{ED}+G(Enc(I))\right)\right) \right], %\label{eq:l_adv_g}
\end{align}
%	Adversarial learning has shown great success in generating photo-realistic images in recent two years~\cite{che2016mode,isola2016image,zhu2017unpaired,kim2017learning,taigman2016unsupervised,pathak2016context}. 
%	However, GANs also suffer from two critical issues, namely, instability and mode missing. Some recent works, e.g., WGAN~\cite{arjovsky2017wasserstein} and LSGAN~\cite{mao2016least}, have stabilized the training of GANs but still cannot ensure the stability of ED+GAN because of the interaction with the reconstruction loss.
%	In ED//GAN, since the adversarial learning is decoupled from the reconstruction learning, any improved GANs could be adopted to specifically ensure stable learning of GAN without affecting or being affected by the ED net. Therefore, there is no need to introduce a weighting factor to trade off the reconstruction and adversarial losses like in ED+GAN. 
%In addition, as demonstrated in~\cite{larsen2015autoencoding,che2016mode}, incorporating GAN with ED can relax the mode missing issue because ED tends to imitate training samples (both major and minor modes). The proposed ED//GAN inherits this property, making GAN generate residual based on the reconstruction (imitation) from ED. In other words, a decoupled ED is able to guide GAN to avoid mode missing. It is also worth noting that by introducing a single/decoupled generator, G, in ED//GAN, direct visualization is possible. This will be elaborated further in Sec.~\ref{subsec:visualization}. 
Finally, we obtain the objective of the proposed decoupled learning (ED//GAN),
\begin{align}
	%\underset{Enc,Dec,G,D}{\min}\mathcal{L}_{const}(Enc,Dec) + \mathcal{L}_{adv}(G) + \mathcal{L}_{adv}(D)\\
	\underset{Enc,Dec}{\min}\mathcal{L}_{const}(Enc,Dec) + \underset{G}{\min} \mathcal{L}_{adv}(G) + \underset{D}{\min}\mathcal{L}_{adv}(D).
	\label{eq:objective}
\end{align}     
Note that there are no weighting parameters between the losses in the objective function, which relaxes the manual tuning that may require an expert with strong domain knowledge and rich experience. 
During training, each component could be updated alternatively and separately because the three components do not overlap in backpropagation, i.e., the backpropagation paths are not intertwined. In practice, however, ED could be trained first because it is completely independent from GAN and GAN operates on the output of ED.

% Assuming ED is parameterized by $\theta$ in both methods, and G in ED//GAN is with the parameter $\phi$, then the update of $\theta$ and $\phi$ through stochastic gradient decent can be expressed by
% \begin{align}
% \text{ED+GAN: } \theta^{t+1} &= \theta^t - \alpha \frac{\partial}{\partial\theta}\left( \mathcal{L}_{const} + \lambda\mathcal{L}_{adv} \right)\\
% ~ & = \theta^t - \alpha \frac{\partial\mathcal{L}_{const}}{\partial\theta} - \alpha\cdot \lambda \frac{\partial\mathcal{L}_{adv}}{\partial\theta} 
% ,\label{eq:grad_couple}\\
% \text{ED//GAN: } \theta^{t+1} &= \theta^t - \alpha \frac{\partial \mathcal{L}_{const} }{\partial\theta},\; \phi^{t+1} \\
% ~ & = \phi^t - \alpha \frac{\partial\mathcal{L}_{adv}}{\partial\phi},\label{eq:grad_decouple}
% \end{align}
% where $\alpha$ denotes the learning rate. In Eq.~\ref{eq:grad_couple}, the loss function in Eq.~\ref{eq:obj_old} (left) is adopted as an example.

\subsection{Boosting Effect from Adversarial Learning}
A side product of the proposed ED//GAN is that it helps to investigate how the discriminator independently boosts the quality of generated images. In ED+GAN, however, the effect of discriminator is difficult to directly identify because it is coupled with the effect of ED.
The learned residual in ED//GAN is considered the boosting factor from the adversarial learning (discriminator). %In other words, the residual illustrates where in the generated images from reconstruction of ED is boosted in pixel level. 
Generally, the images from ED tend to be blurry, while the residual from GAN carries the details or important texture information for photo-realistic image generation. Imposing the residual onto the reconstructed images is supposed to yield higher-fidelity images as compared to the reconstructed images.

\begin{figure}[ht]
	\centering
	\includegraphics[width=.9\columnwidth]{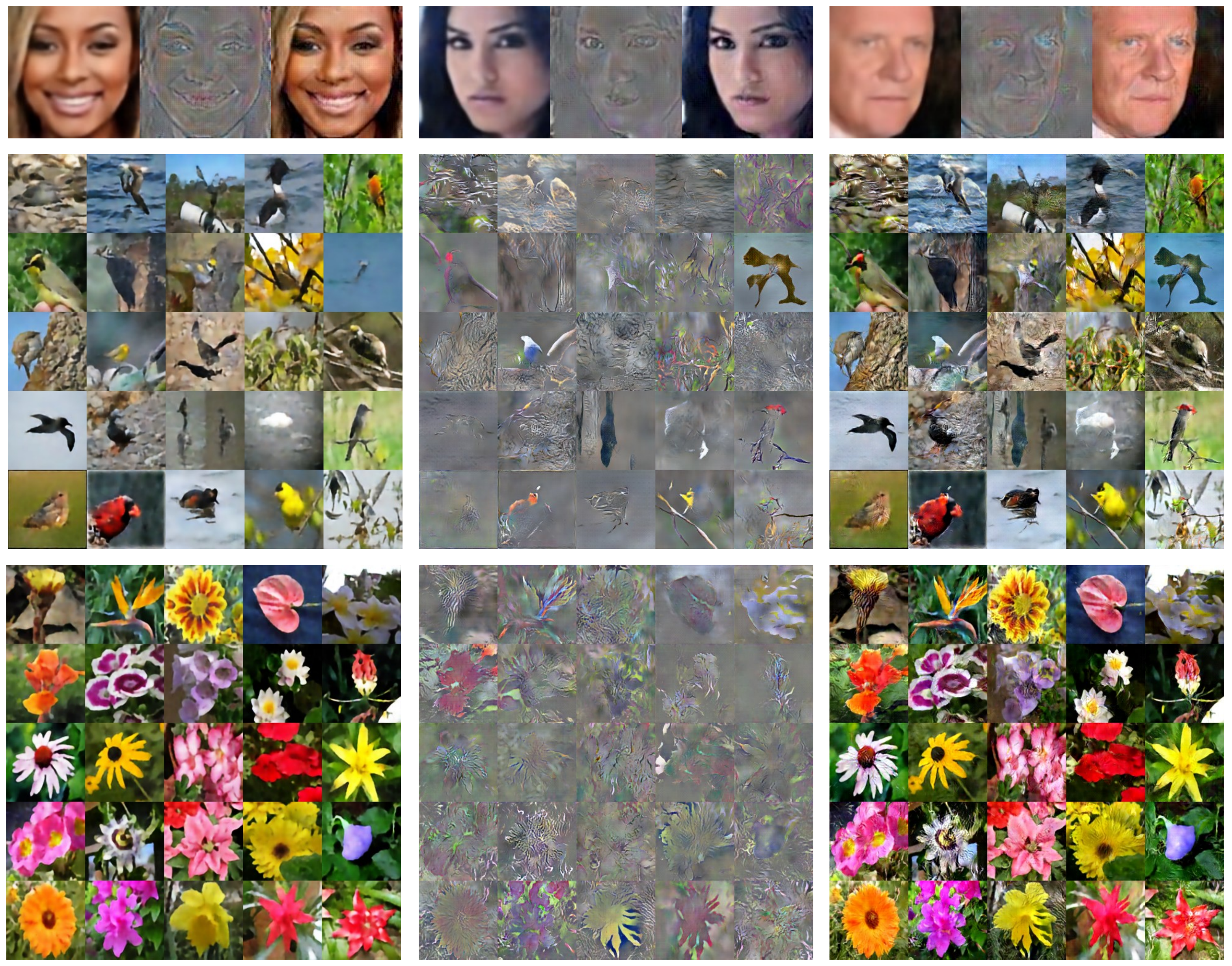}
	\caption{Visualization of the boost from adversarial learning trained on UTKFace~\cite{zhang2017age}, CUB-200~\cite{wah2011caltech}, and Oxford Flower~\cite{Nilsback08} datasets. From left to right in each triple: reconstruction, residual, and output images from ED//GAN.}
	\label{fig:visual}
\end{figure}

In Fig.~\ref{fig:visual} (middle picture in each triple), we can observe that the adversarial learning mainly enhances the edges at eyebrow, eyes, mouth, and teeth for face images. For the bird and flower images, the residues further enhance the color. In some cases, the added details also create artifacts. In general, adding the residual to the blurry images from ED (Fig.~\ref{fig:visual} left), the output images present finer details. 
%Fig.~\ref{fig:visual} illustrates the learning results from the proposed ED//GAN. The examples are randomly selected with different age, race, expression, and pose. 

An argument on the visualization of adversarial effect may be that subtracting the result of ED from that of ED+GAN could also obtain the residual. Although this process can roughly visualize the boost from GAN, we emphasize that ``ED+GAN'' minus ``ED'' is not purely the effect from GAN because the training of GAN is affected by ED in ED+GAN and vice versa. In the proposed ED//GAN, however, ED is trained independently from GAN, thus GAN only learns the residual between real images and those from ED. %in a straightforward manner.    

\section{Normalized Relative Discriminative Score}
In the evaluation of image quality (e.g., reality and resolution), how to design a reliable metric for generative models has been an open issue. Existing metrics (e.g., inception score~\cite{salimans2016improved} and related methods~\cite{odena2016conditional}), although successful in certain cases, have been demonstrated to be problematic in others~\cite{metz2016unrolled}. If a perfect metric exists, the training of generative models would be much easier because we could use such metric as loss directly without training a discriminator. 
The rationale behind our design is that if it is difficult to obtain the absolute score (perfect metric) of a model, we could at least compare which model generates better images than others. 
From this perspective, we propose to perform relative comparison rather than providing evaluation based on absolute score like existing works. More specifically, we train a single discriminator/classifier to separate real samples from generated samples, and those generated samples closer to real ones will be more difficult to be separated. For example, given two generative models G$_1$ and G$_2$, which define the distributions of generated samples $p_{g1}$ and $p_{g2}$, respectively. Suppose the distribution of real data is $p_{data}$, if $JSD(p_{g1}|p_{data}) < JSD(p_{g2}|p_{data})$ where $JSD$ denotes the Jensen-Shannon divergence and assume $p_{g1}$ and $p_{g2}$ intersect with $p_{data}$, a discriminator/classifier D trained to classify real samples as 1 and 0 otherwise would show the following inequality,
\begin{equation}
	\mathbb{E}_{x\sim p_{data}}[D(x)] \ge \mathbb{E}_{x\sim p_{g1}}[D(x)] \ge \mathbb{E}_{x\sim p_{g2}}[D(x)].
\end{equation}

\begin{figure}[ht]
	\centering
	\includegraphics[width=.7\columnwidth]{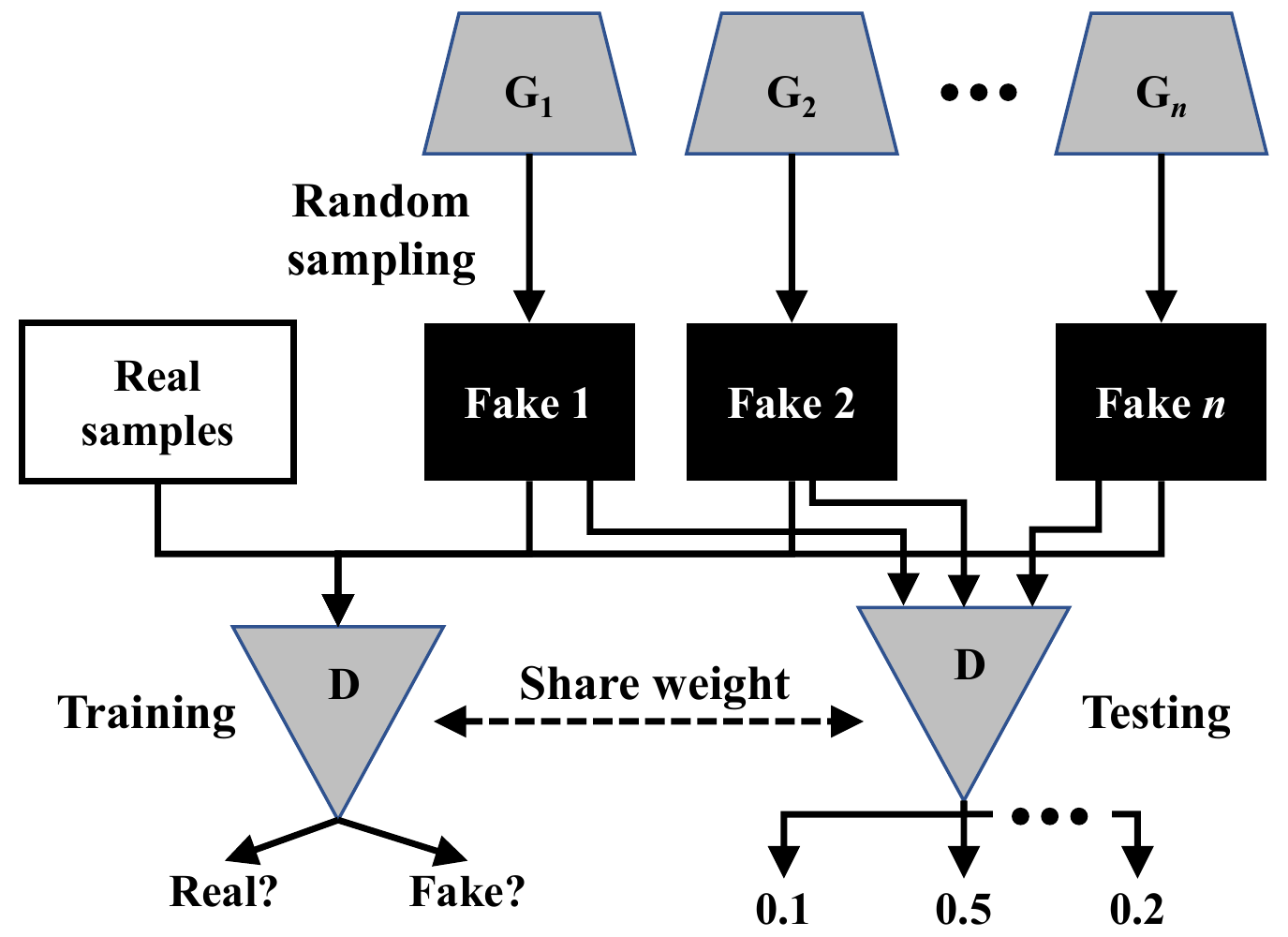}
	\caption{Illustration of NRDS. G$_n$ indicates the $n$th generative model, and its corresponding fake samples are Fake $n$, which are smapled randomly. The fake samples from $n$ models, as well as the real samples, are used to train the binary classifier D (bottom left). Testing only uses fake samples and performs alternatively with the traing process. The bottom right shows an example of averaged output of D from fake samples of each model.}
	\label{fig:nrds}
\end{figure}

The main idea is that if the generated samples are closer to real ones, more epochs would be needed to distinguish them from real samples. The discriminator is a binary classifier to separate the real samples from fake ones generated by all the models in comparison. In each epoch, the discriminator output of each sample is recorded. The average discriminator output of real samples will increase with epoch (approaching 1), while that of generated samples from each model will decrease with epoch (approaching 0). However, the decrement rate of each model varies based on how close the generated samples to the real ones. Generally, the samples closer to real ones show slower decrement rate. Therefore, we compare the ``decrement rate'' of each model to relatively evaluate their generated images. The decrement rate is proportional to the area under the curve of average discriminator output versus epoch. Larger area indicates slower decrement rate, implying that the generated samples are closer to real ones. Fig.~\ref{fig:nrds} illustrates the computation of normalized relative discriminative score (NRDS).

There are three steps to compute the proposed normalized relative discriminative score (NRDS): 1) Obtain the curve $\mathcal{C}_i$ ($i=1,2,\cdots,n$) of discriminator output versus epoch (or mini-batch) for each model (assuming $n$ models in comparison) during training; 2) Compute the area under each curve $A(\mathcal{C}_i)$; and 3) Compute NRDS of the $i$th model by 
\begin{equation}
	NRDS_i = \frac{A(\mathcal{C}_i)}{\sum_{j=1}^{n}A(\mathcal{C}_j)}.
	\label{eq:nrds}
\end{equation}
%$NRDS_i = \frac{A(\mathcal{C}_i)}{\sum_{j=1}^{n}A(\mathcal{C}_j)}$.

\begin{figure}[ht]
	\centering
	\includegraphics[width=.5\columnwidth]{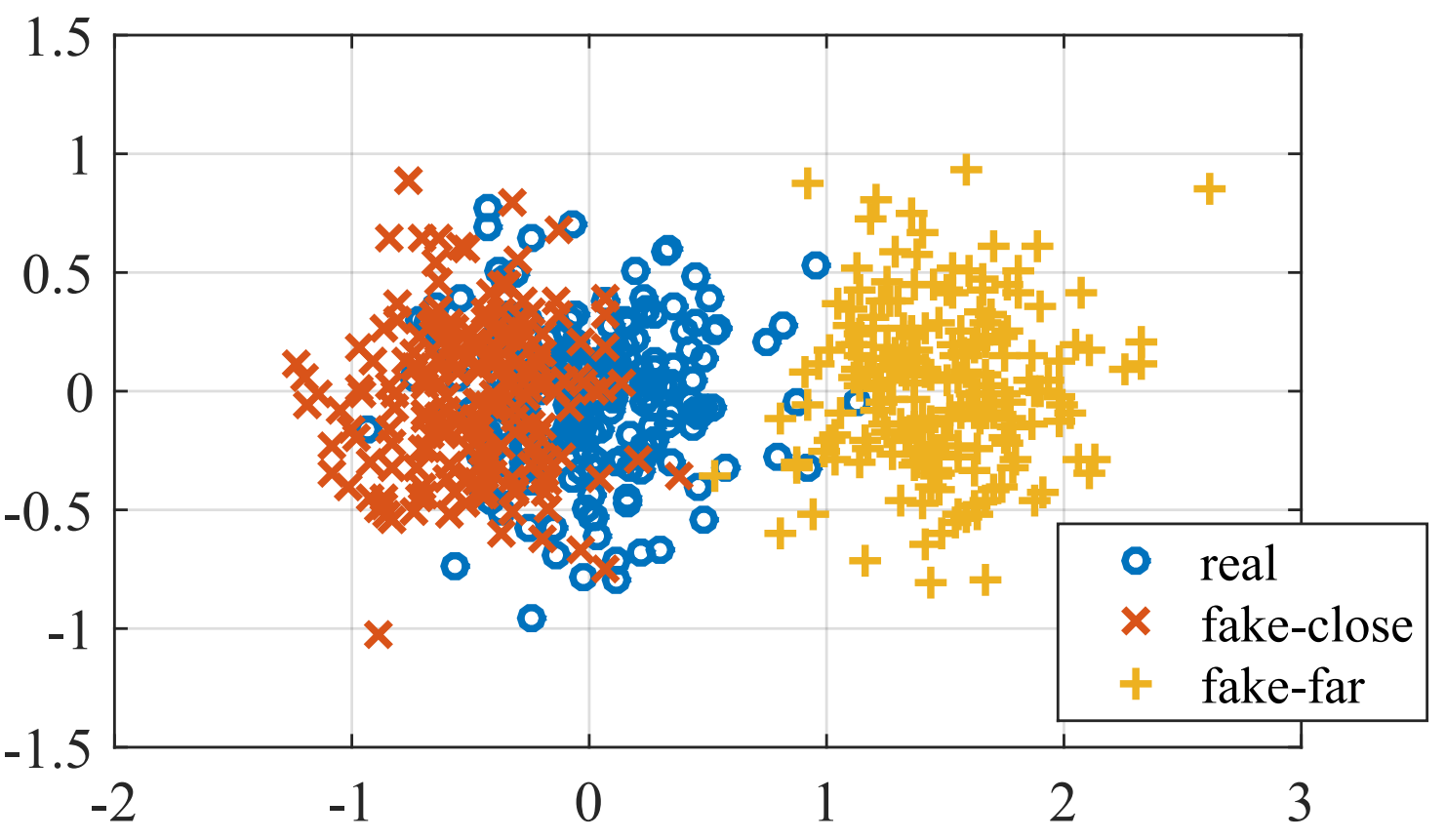}
	\includegraphics[width=.4\columnwidth]{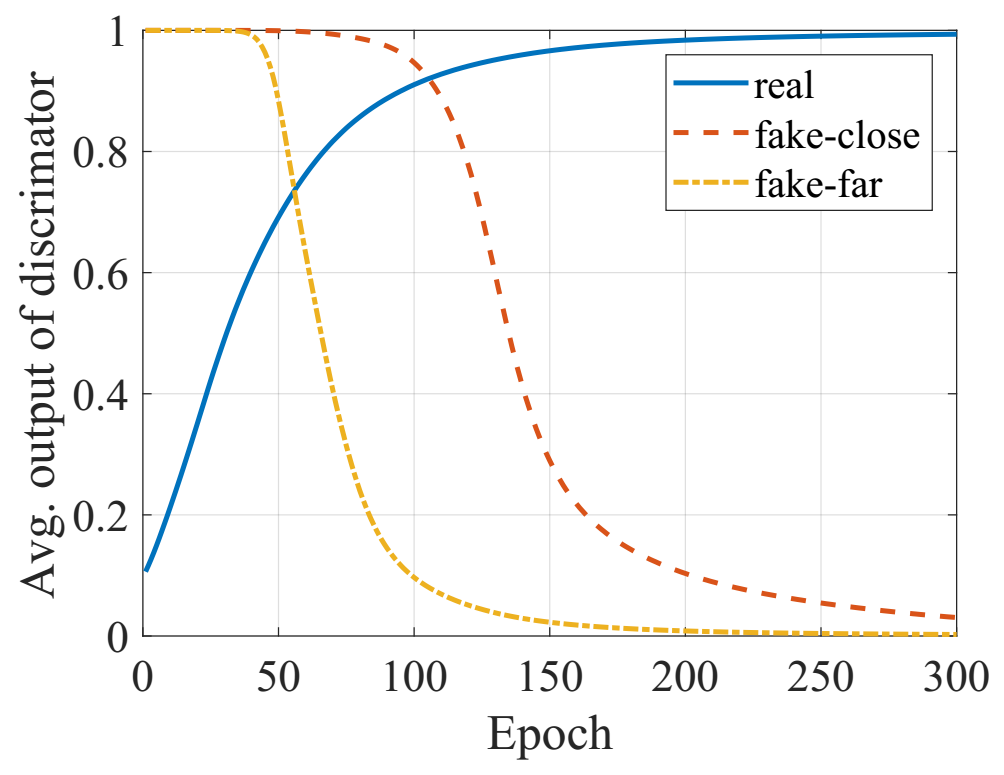}
	%\caption{A toy example of computing NRDS. \textbf{Left:} the real and fake samples. \textbf{Right:} the curves of epoch vs. averaged output of discriminator.}
	\caption{ A toy example of computing NRDS. Left: the real and fake samples randomly sampled from 2-D normal distributions with different means but with the same (identity) covariance. The real samples (blue circle) is with zero mean. The red ``x'' and yellow ``+'' denote fake samples with the mean of $[−0.5, 0]$ and $[1.5, 0]$, respectively. The notation fake-close/far indicates that
		the mean of correspondingly fake samples is close to or far from that of the real samples. Right: the curves of epoch vs. averaged output of discriminator on corresponding sets (colors) of samples.}
	\label{fig:nrds_demo}
\end{figure}

To illustrate the computation of NRDS, Fig.~\ref{fig:nrds_demo} shows a toy example. Assume the samples named ``fake-close'' and ``fake-far'' are generated from two different models to simulate the real samples. We train a discriminator on the real and fake (i.e., fake-close and fake-far) samples. The structure of discriminator is a neural network with two hidden layers, both of which have 32 nodes, and ReLU is adopted as the activation function. 
After each epoch of training on the real and fake samples, the discriminator is tested on the same samples from real, fake-close, and fake-far, respectively. For example, all the real samples are fed to the discriminator, and then we compute the mean of the outputs from the discriminator. By the same token, we can obtain the average outputs of fake-close and fake-far, respectively. With 300 epochs, we plot the curves shown in Fig.~\ref{fig:nrds_demo} (right). Intuitively, the curve of fake-close approaches zero slower than that of fake-far because the samples in fake-close are closer (similar) to the real samples. 
The area under the curves of fake-close ($\mathcal{C}_1$) and fake-far ($\mathcal{C}_2$) are $A(\mathcal{C}_1)=145.4955$ and $A(\mathcal{C}_2)=71.1057$, respectively. From Eq.~\ref{eq:nrds}, 
\begin{align}
	NRDS_1 &=\frac{A(\mathcal{C}_1)}{\sum_{i=1}^{2}A(\mathcal{C}_i)}=0.6717\\
	NRDS_2 &=\frac{A(\mathcal{C}_2)}{\sum_{i=1}^{2}A(\mathcal{C}_i)}=0.3283.
\end{align}
%$NRDS_1=0.6717$ and $NRDS_2=0.3283$. 
Therefore, we can claim that the model generating fake-close is relatively better. 
Note that the actual value of NRDS for certain single model is meaningless. We can only conclude that the model with higher NRDS is better than those with lower NRDS in the same comparison, but a high NRDS does not necessarily indicate an absolutely good model.  

\section{Experimental Evaluation}
\label{sec:experimant}

We evaluate the proposed decoupled learning mainly from 1) its ability in relaxing the weight setting and 2) its generality in adapting to existing works. First, we compare the proposed ED//GAN to the traditional ED+GAN based on the UTKFace dataset~\cite{zhang2017age} using a general (not fine-tuned) network structure. Then, two existing works, i.e., Pix2Pix~\cite{isola2016image} and CAAE~\cite{zhang2017age}, are adopted for adaptation, where the corresponding datasets are use, i.e. UTKFace and CMP Facade databases~\cite{Tylecek13}, respectively. 

The UTKFace dataset consists of about 20,000 aligned and cropped faces with large diversity in age and race. The decoupled learning applied on the UTKFace dataset aims to demonstrate the performance on image manipulation tasks. The CMP Facade dataset is utilized to illustrate the performance of the decoupled learning on image transformation tasks. %The experimental results validate the robustness and stability of the decoupled learning in parameter relaxation, i.e., guaranteeing stable training
without parameter tuning on any datasets.

\subsection{Comparison between ED//GAN and ED+GAN}
\label{subsec:exp_train}	
For fair comparison, we compare ED//GAN and ED+GAN on the same network and dataset. \textit{This network is neither specifically designed for any applications nor delicately fine-tuned to achieve the best result. The goal is to illustrate the advantages of ED//GAN as compared to ED+GAN}. Table~\ref{tab:nets} details the network structure used in this experiment.
\begin{table}[ht]
	\centering
	\caption{The network structure shown in Fig.~\ref{fig:flow}. The size of each layer is denoted by $h\times w\times c$, corresponding to height, width, and number of channels, respectively.}
	\begin{tabular}{l|l}
		\hline
		Enc / D & Size \\
		\hline
		Input & $128\times 128 \times 3$ \\ 
		Conv, BN, ReLU& $64 \times 64 \times 64$ \\
		Conv, BN, ReLU & $32 \times 32 \times 128$ \\
		Conv, BN, ReLU & $16 \times 16 \times 256$ \\
		Conv, BN, ReLU & $8 \times 8 \times 512$ \\
		Conv, BN, ReLU & $4 \times 4 \times 1024$ \\
		Reshape, FC, tanh & $50$ / $1$ \\
		\hline\hline
		Dec / G & Size\\
		\hline
		Input & 50 \\
		FC, ReLU, BN, Reshape & $4\times 4 \times 1024$ \\
		Deconv, BN, ReLU & $ 8\times 8 \times 512$ \\
		Deconv, BN, ReLU & $ 16\times 16\times 256$ \\
		Deconv, BN, ReLU & $ 32\times 32\times 128$ \\
		Deconv, BN, ReLU& $ 64\times 64\times 64$ \\
		Deconv, tanh & $ 128\times 128\times 3$ \\
		\hline
	\end{tabular}
	\label{tab:nets}
\end{table} 

\begin{figure}[ht]
	\centering
	\includegraphics[width=\columnwidth]{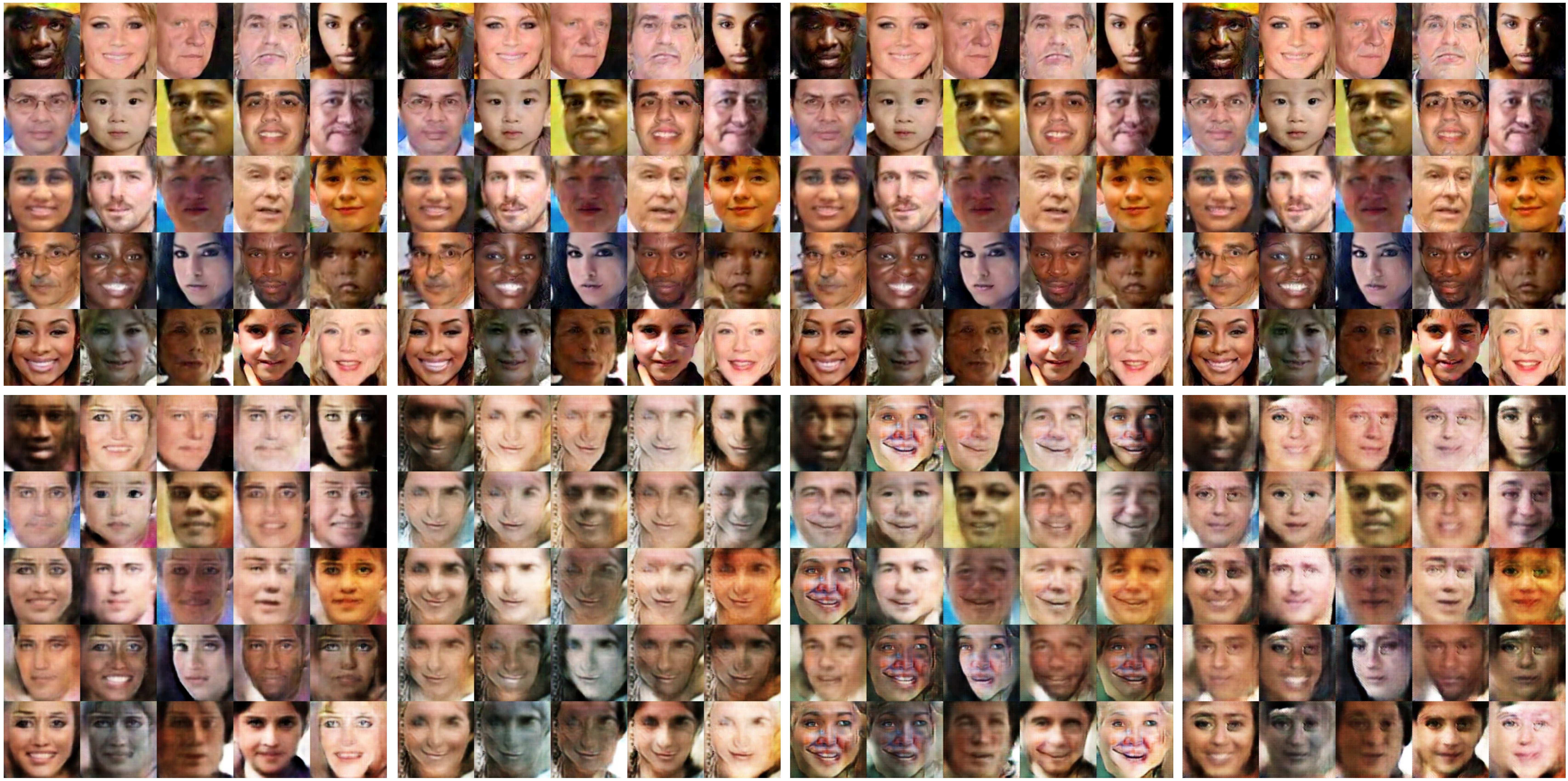}
	\caption{Comparison of ED//GAN (top) and ED+GAN (bottom) on the UTKFace dataset. From left to right, the weights on the adversarial loss are 0.001, 0.01, 0.1, and 1, respectively. Please zoom in for better view.}
	\label{fig:cmp_bn}
\end{figure}

To demonstrate the effectiveness of ED//GAN on relaxing the weight setting, we compare it to ED+GAN by changing the weight between the reconstruction loss and adversarial loss. In the objective function of ED//GAN (Eq.~\ref{eq:objective}), no weight is required. For comparison purpose, we intentionally add a weighting parameter $\lambda$ to the adversarial loss like the objective of GAN (Eq.~\ref{eq:obj_old}). Iterating $\lambda$ from 0.001 to 1 with the step of 10x, we obtain the results as shown in Fig.~\ref{fig:cmp_bn} after 200 epochs with the batch size of 25.
The output images from ED//GAN are relatively higher-fidelity and maintain almost the same quality regardless of the weight change. However, the outputs of ED+GAN significantly vary with the weight. In addition, ED+GAN generates unstable results, e.g., model collapsing and significant artifacts. The corresponding NRDS is calculated in Table~\ref{tab:nrds1}, where the number of models in comparison is $i=8$ (Eq.~\ref{eq:nrds}). The discriminator adopted in NRDS is the same as D in Table~\ref{tab:nets}. 
\begin{table}[ht]                       
	\centering
	\caption{NRDS for different weight settings (Fig.~\ref{fig:cmp_bn}).}
	\begin{tabular}{c|c|c|c|c|c}
		\hline
		~ & 0.001 & 0.01 & 0.1 & 1 & std \\
		\hline
		ED+GAN  & .1066 & .1143 & .1268 & .1267& .0099\\
		ED//GAN &  .1320 & .1300 & .1300 & .1336 &.0017\\
		\hline
	\end{tabular}
	\label{tab:nrds1}
\end{table}

\begin{figure}[ht]
	\centering
	\includegraphics[width=\columnwidth]{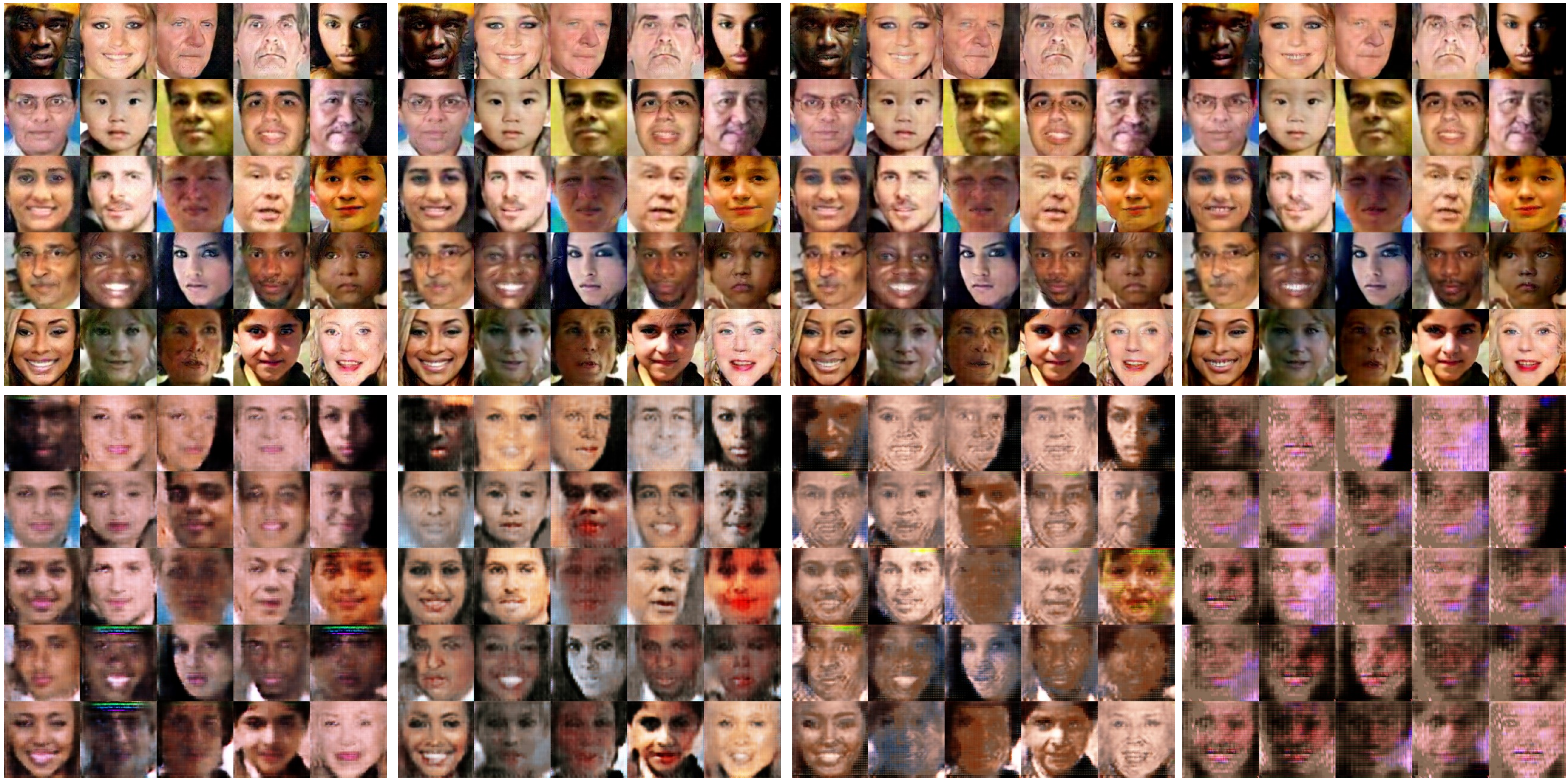}
	\caption{Comparison between ED//GAN (top) and ED+GAN (bottom) without batch normalization on the UTKFace dataset. From left to right, the weights on the adversarial loss are 0.001, 0.01, 0.1, and 1, respectively.}
	\label{fig:cmp_nobn}
\end{figure}

Now, we remove the batch normalization in Enc and Dec to see whether ED//GAN still yields stable results. Fig.~\ref{fig:cmp_nobn} compares the results from ED//GAN and ED+GAN by removing the batch normalization in Enc and Dec. The corresponding NRDS is listed in Table~\ref{tab:nrds2}.
\begin{table}[ht]                       
	\centering
	\caption{NRDS for different weight settings (Fig.~\ref{fig:cmp_nobn}).}
	\begin{tabular}{c|c|c|c|c|c}
		\hline
		~ & 0.001 & 0.01 & 0.1 & 1 & std \\
		\hline
		ED+GAN & .1172 & .1143 & .1163 & .0731 & .0215 \\
		ED//GAN & .1432 & .1434 & .1458 & .1466 &.0017 \\
		\hline
	\end{tabular}
	\label{tab:nrds2}
\end{table}

From the two experiments, ED//GAN vs. ED+GAN with/without batch normalization on ED (i.e., Enc and Dec), we observe that ED//GAN generally yields higher NRDS, indicating better image quality. In addition, the NRDS values for ED//GAN vary much less than those of ED+GAN, as observed from the lower standard deviation (std), indicating robustness against different weights. These observations completely agree with our claim --- ED//GAN stabilizes the training regardless of the trade-off issue in the traditional ED+GAN structure. %Therefore, the proposed decoupled learning (ED//GAN) could relax the weight setting.

We notice that for ED//GAN, the NRDS value slightly changes with the change of weight. However, the change is too small to be observable from visual inspection. We also observe that NRDS achieves the peak value at certain weight settings. For example, NRDS achieves the highest value at $\lambda=1$ in both Tables~\ref{tab:nrds1} and \ref{tab:nrds2}, which happens to be the case of the proposed ED//GAN without weight setting. %We also point out that although such changes can only be observed from the NRDS score, rather than from visual perception.  which emphasizes our claim that ED//GAN relaxes the weight setting dilemma in ED+GAN.

\subsection{Adaptation from Existing Works to ED//GAN}
An essential merit of ED//GAN is its adaptability for existing ED+GAN works. Specifically, an existing work that adopted the ED+GAN structure could be easily modified to the ED//GAN structure without significantly reducing the performance but with the benefit of not having to fine-tune the weight. To demonstrate the adaptability of ED//GAN, we modify  two existing works: 1) Pix2Pix~\cite{isola2016image} for image transformation and 2) CAAE~\cite{zhang2017age} for image manipulation. According to Fig.~\ref{fig:demo}, the modification is simply to parallelize a G (the copy of Dec) to the original network. The objective functions are modified from Eq.~\ref{eq:obj_old} to Eq.~\ref{eq:obj_new}, which is straightforward to implement.

\subsubsection{Adaptation on Pix2Pix}
\begin{figure}[ht]
	\centering
	\includegraphics[width=\columnwidth]{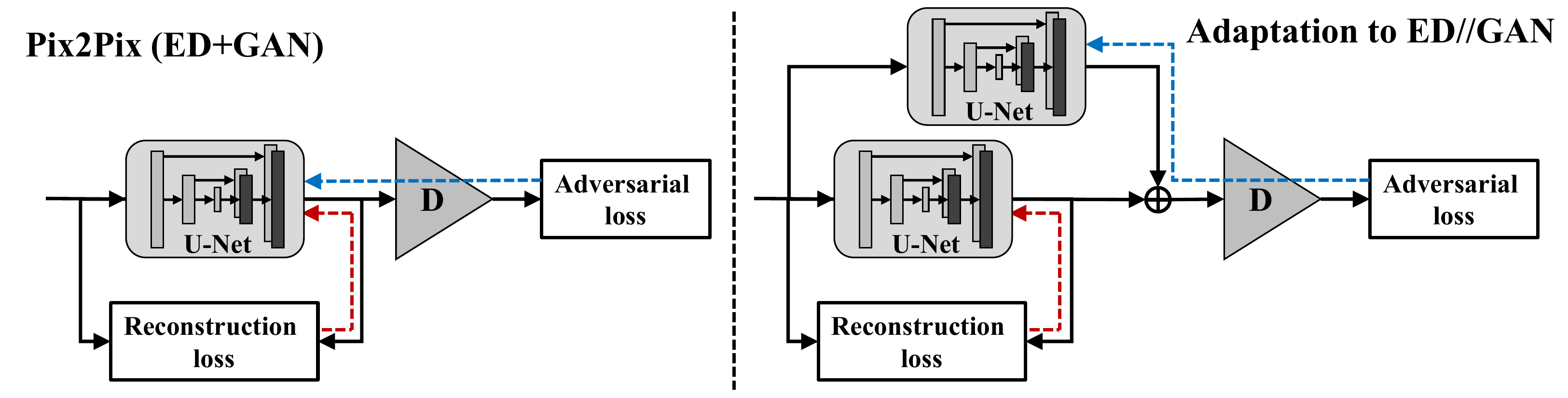}
	\caption{Left: the network structure of Pix2Pix (ED+GAN). Right: the adaptation to the proposed ED//GAN. Solid black arrows denote the feedforward path, and dashed arrows in red and blue indicate backpropagation from the reconstruction loss and the adversarial loss, respectively.}
	\label{fig:pix2pix}
\end{figure}

We adapt the network in Pix2Pix~\cite{isola2016image}, which is ED+GAN structure, to the proposed ED//GAN structure as shown in Fig.~\ref{fig:pix2pix}. 

In Pix2Pix, ED is implemented by the U-Net, which directly passes feature maps from encoder to decoder, preserving more details. In order not to break the structure of U-Net, we apply another U-Net as the generator G in the corresponding ED//GAN version. Fig.~\ref{fig:p2p} compares the results from Pix2Pix and its ED//GAN version. The reported weight in Pix2Pix is 100:1, where the weight on reconstruction loss is 100, and 1 on the adversarial loss. We change the weight setting to 1:1 and 1000:1 to illustrate its effect on the generated images.
\newcommand{\wt}{.11\columnwidth}
\begin{figure}[ht]
	\centering
	\begin{tabular}{p{\wt}p{\wt}p{\wt}p{\wt}p{\wt}p{\wt}}%{cccccc}
		\scriptsize\centering Input & \scriptsize\centering Real & \scriptsize\centering 1:1 & \scriptsize\centering 100:1 & \scriptsize\centering 1000:1 & \scriptsize\centering ED//GAN
	\end{tabular}\\
	\includegraphics[width=\columnwidth]{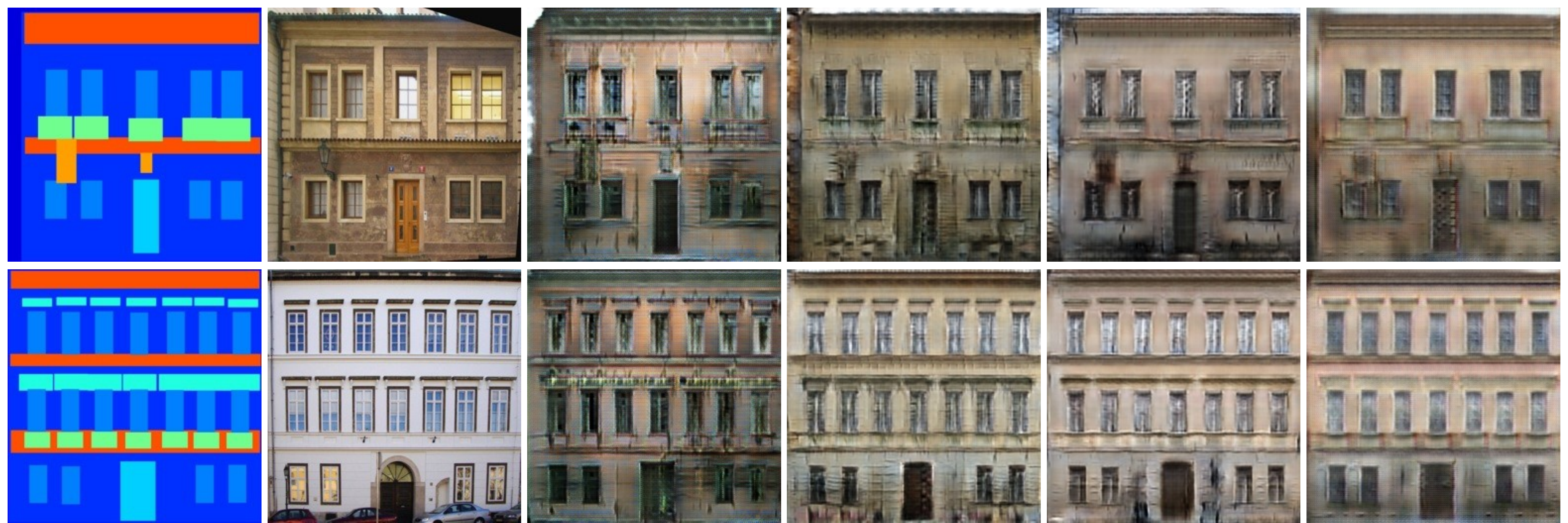}
	\caption{Comparison between Pix2Pix and its ED//GAN version. Pix2Pix generates images at different weight settings as denoted by $\lambda$:1, where $\lambda$ and 1 indicate the weights of the reconstruction loss and adversarial loss, respectively. ED//GAN denotes the generated images from the modified decoupled structure.
	}
	\label{fig:p2p}
\end{figure}

We observe that the generated images with the weight of 1:1 introduce significant artifacts (zoom in for better view). With higher weight on the reconstruction loss, e.g., 100:1 and 1000:1, more realistic images can be generated, whose quality is similar to that from the ED//GAN version that does not need weight setting.
\newcommand{\wtt}{.4\columnwidth}
\begin{figure*}[ht]
	\centering
	\begin{tabular}{p{\wtt}p{\wtt}p{\wtt}p{\wtt}}
		\small\centering 1:$10^{-4}$ & \small\centering 1:$10^{-2}$ & \small\centering 1:1 & \small\centering ED//GAN
	\end{tabular}\\
	\includegraphics[width=1.8\columnwidth]{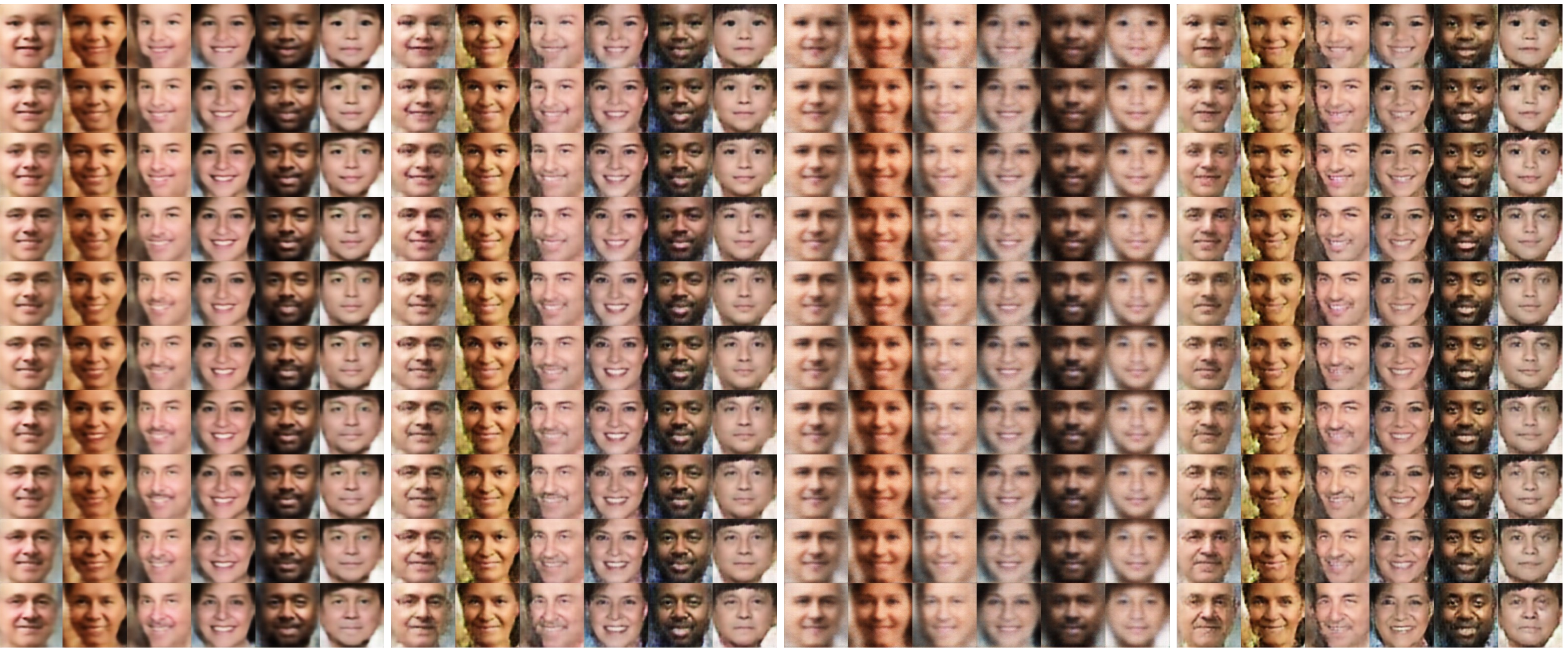}
	\caption{Comparison between CAAE~\cite{zhang2017age} and its ED//GAN version. CAAE generates images at different weights as denoted by 1:$\lambda$, where 1 and $\lambda$ indicate the weights of the reconstruction loss and adversarial loss, respectively. ED//GAN denotes the generated images from the modified decoupled structure.
	}
	\label{fig:caae}
\end{figure*}
\subsubsection{Adaptation on CAAE}  
\begin{figure}[ht]
	\centering
	\includegraphics[width=\columnwidth]{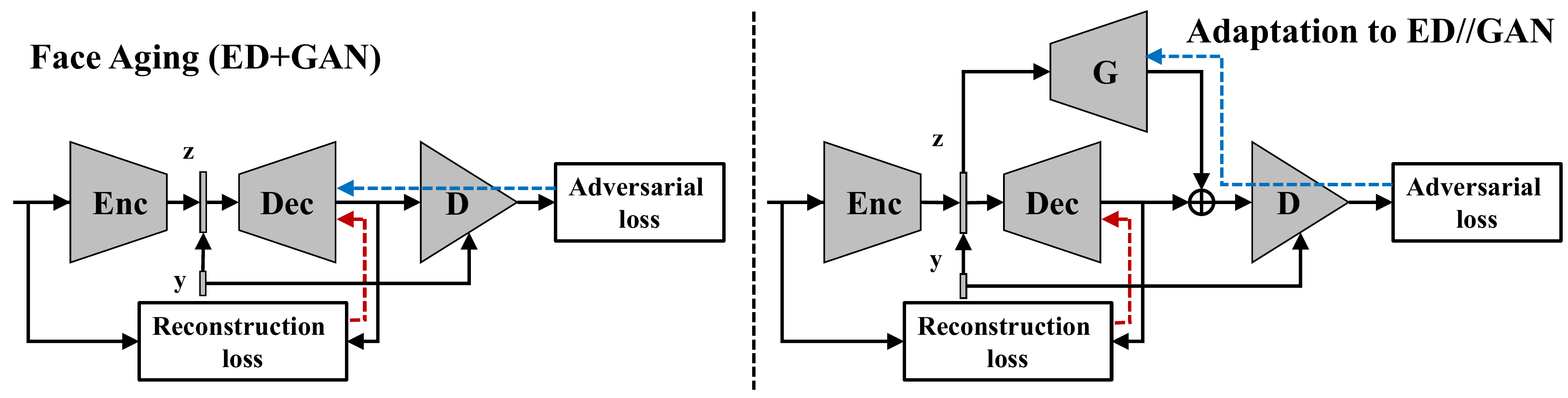}
	\caption{Left: the ED+GAN structure used in CAAE~\cite{zhang2017age} (the discriminator on $z$ is neglected for simplicity because the modification will not affect that part). Right: the adaptation to the proposed ED//GAN. Solid black arrows denote the feedforward path, and dashed arrows in red and blue indicate backpropagation from the reconstruction loss and the adversarial loss, respectively. The label $y$ is concatenated to $z$ and D to control the age.}
	\label{fig:caae2}
\end{figure}

We next adapt CAAE~\cite{zhang2017age}, a conditional ED+GAN structure, to the proposed ED//GAN structure as shown in Fig.~\ref{fig:caae2}. CAAE generates aged face by manipulating the label concatenated to the latent variable $z$ from Enc.  

The original network used in CAAE has an extra discriminator on $z$ to force $z$ to be uniformly distributed. We do not show this discriminator in Fig.~\ref{fig:caae2} because it does not affect the adaptation. 
Fig.~\ref{fig:caae} shows some random examples to compare the original and modified structures. The weights of the reconstruction loss and adversarial loss are 1 and $10^{-4}$ (i.e., 1:$10^{-4}$) as reported in the original work. We use another two different weight settings, 1:$10^{-2}$ and 1:1, for the original structure and compare the results with the corresponding ED//GAN version. 

The NRDS is provided for both adaptation experiments to statistically analyze the adaptability of ED//GAN, as shown in Tables~\ref{tab:nrds_p2p} and \ref{tab:nrds_caae}, respectively. 
\begin{table}[ht]
	\centering
	\caption{NRDS for Pix2Pix adaptation.}
	\begin{tabular}{c|c|c|c|c}
		\cline{1-5}
		\multirow{2}{*}{Method} & \multicolumn{3}{|c|}{ED+GAN} & ED//GAN \\ 
		\cline{2-4}
		~ & 1:1 & 100:1 & 1000:1 \\
		\cline{1-5}
		NRDS & .2190 & \textbf{.2641} & .2572 & .2597 \\
		\cline{1-5}
	\end{tabular}
	\label{tab:nrds_p2p}\\
	\begin{tabular}{c}
		~
	\end{tabular}\\
	\caption{NRDS for CAAE adaptation.}
	\begin{tabular}{c|c|c|c|c}
		\cline{1-5}
		\multirow{2}{*}{Method} & \multicolumn{3}{|c|}{ED+GAN} & ED//GAN  \\ 
		\cline{2-4}
		~ & 1:$10^{-4}$ & 1:$10^{-2}$ & 1:1 & ~ \\
		\cline{1-5}
		NRDS & .2527 & .2496 & .2430 &  \textbf{.2547} \\
		\cline{1-5}
	\end{tabular}
	\label{tab:nrds_caae}
\end{table}
We observe that the ED//GAN structure still, in general, yields higher or similar NRDS values than the coupled counterpart. Although in Table~\ref{tab:nrds_p2p}, the proposed ED//GAN ranks number two, ED//GAN achieves the competitive result without the need of tuning the weight parameter. In Table~\ref{tab:nrds_caae}, ED//GAN ranks the top as compared to the ED+GAN structure. Note that we show alternatives of parameter setting based on the optimal settings that are already known from the original papers. If designing a new structure without any prior knowledge, however, it could be difficult to find out the optimal weight with only a few trials. %Therefore, we are comparing to the peak performance of the two existing works.

It is worth emphasizing that the goal is not to beat the best result from fine-tuned ED+GAN. Rather, ED//GAN aims at achieving stable and competitive results without having to fine-tune the weight.% such that the performance of the network structure is stable and If performing exhaustive search for the weight between two losses in ED+GAN, the NRDS may drastically increase or decrease and achieve the peak value at certain point. By contrast, ED//GAN will keep the similar NRDS close to the peak value. Therefore, compared to the ED+GAN structure, ED//GAN is with higher confidence to ensure appealing results.       

\section{Conclusion}
This paper proposed the novel decoupled learning structure (ED//GAN) for image generation tasks with image-conditional models. Different from existing works where the reconstruction loss (from ED) and the adversarial loss (from GAN) are backpropagated to a single decoder, referred to as the coupled structure (ED+GAN), in ED//GAN, the two losses are backpropagated through separate networks, thus avoiding the interaction between each other. The essential benefit of the decoupled structure is such that the weighting factor that has to be fine-tuned in ED+GAN is no longer needed in the decoupled structure, thus improving stability without looking for the best weight setting. This would largely facilitate the wider realization of more specific image generation tasks. The experimental results demonstrated the effectiveness of the decoupled learning. We also showed that existing ED+GAN works can be conveniently modified to ED//GAN by adding a generator that learns the residual. %The ED//GAN could ensure stable and competitive results without parameter tuning.

\newpage
%{\small
\bibliographystyle{ieee}
\bibliography{references}
%}

\end{document}